%% file: root.tex
\documentclass[letterpaper, 10 pt, conference]{ieeeconf}  

\IEEEoverridecommandlockouts                              
\usepackage[utf8]{inputenc}
\usepackage[T1]{fontenc}

\usepackage{graphicx} 
\graphicspath{{./img/}}
\usepackage{amsmath} 
\usepackage{amssymb}  
\usepackage{tikz, pgfplots}
\usetikzlibrary{svg.path}
\usepackage{scalerel}

\definecolor{orcidlogocol}{HTML}{A6CE39}
\tikzset{
  orcidlogo/.pic={
    \fill[orcidlogocol] svg{M256,128c0,70.7-57.3,128-128,128C57.3,256,0,198.7,0,128C0,57.3,57.3,0,128,0C198.7,0,256,57.3,256,128z};
    \fill[white] svg{M86.3,186.2H70.9V79.1h15.4v48.4V186.2z}
                 svg{M108.9,79.1h41.6c39.6,0,57,28.3,57,53.6c0,27.5-21.5,53.6-56.8,53.6h-41.8V79.1z M124.3,172.4h24.5c34.9,0,42.9-26.5,42.9-39.7c0-21.5-13.7-39.7-43.7-39.7h-23.7V172.4z}
                 svg{M88.7,56.8c0,5.5-4.5,10.1-10.1,10.1c-5.6,0-10.1-4.6-10.1-10.1c0-5.6,4.5-10.1,10.1-10.1C84.2,46.7,88.7,51.3,88.7,56.8z};
  }
}

\newcommand\orcidicon[1]{\href{https://orcid.org/#1}{\mbox{\scalerel*{
\begin{tikzpicture}[yscale=-1,transform shape]
\pic{orcidlogo};
\end{tikzpicture}
}{|}}}}

\usepackage{hyperref} 

\newcommand\copyrighttext{%
    \scriptsize \copyright{ }2021 IEEE. Personal use of this material is permitted. Permission from IEEE must be obtained for all other uses, in any current or future media, including reprinting/republishing this material for advertising or promotional purposes, creating new collective works, for resale or redistribution to servers or lists, or reuse of any copyrighted component of this work in other works.}
\newcommand\copyrightnotice{%
    \begin{tikzpicture}[remember picture,overlay]
    \node[anchor=south,yshift=10pt,xshift=7pt] at (current page.south) {\parbox{\dimexpr\textwidth-\fboxsep-\fboxrule\relax}{\copyrighttext}};
    \end{tikzpicture}%
}

\title{\LARGE \bf
A Simulation-based End-to-End Learning Framework \\for Evidential Occupancy Grid Mapping*
}

\author{Raphael van Kempen\,\textsuperscript{\orcidicon{0000-0001-5017-7494}}\,, Bastian Lampe\,\textsuperscript{\orcidicon{0000-0002-4414-6947}}\,, Timo Woopen\,\textsuperscript{\orcidicon{0000-0002-7177-181X}}\,, and Lutz Eckstein
\thanks{*This research is accomplished within the project ”UNICARagil” (FKZ 16EMO0289). We acknowledge the financial support for the project by the Federal Ministry of Education and Research of Germany (BMBF).}
\thanks{The authors are with the Institute for Automotive Engineering (ika), RWTH Aachen University, 52074 Aachen, Germany
        {\tt\small \{firstname.lastname\}@ika.rwth-aachen.de}}%
}

\begin{document}
\bstctlcite{IEEEexample:BSTcontrol}

\maketitle

\thispagestyle{empty}
\pagestyle{empty}

\copyrightnotice

\begin{abstract}
Evidential occupancy grid maps (OGMs) are a popular representation of the environment of automated vehicles. Inverse sensor models (ISMs) are used to compute OGMs from sensor data such as lidar point clouds. Geometric ISMs show a limited performance when estimating states in unobserved but inferable areas and have difficulties dealing with ambiguous input. Deep learning-based ISMs face the challenge of limited training data and they often cannot handle uncertainty quantification yet. We propose a deep learning-based framework for learning an OGM algorithm which is both capable of quantifying first- and second-order uncertainty and which does not rely on manually labeled data. Results on synthetic and on real-world data show superiority over other approaches. Source code and datasets are available at \url{https://github.com/ika-rwth-aachen/EviLOG}.
\end{abstract}


\section{INTRODUCTION}

Automated vehicles rely on an accurate model of their environment for planning safe and efficient behavior. Depending on the chosen representation of the environment in this model, different perception algorithms and different sensor modalities are best suited, each coming with corresponding advantages and disadvantages. Often, several different approaches are combined to compensate for the disadvantages of one perception algorithm with the advantages of another.

One common representation of the dynamic environment are object lists. They contain the state of all objects detected and tracked by the vehicle. Several methods for detection and tracking of objects in camera, radar and lidar data have been published during the past years \cite{Aeberhard.2017, Zhou.2018, Lang.2019b}. A drawback of object lists is that the perception algorithms generating them can often only detect a fixed set of predefined object classes. Objects of these classes need to be explicitly contained in the perception algorithms' training data. Since there exist extremely many classes of objects in the world, it is difficult to ensure that all relevant objects are accounted for. 

Occupancy grid mapping algorithms, which are agnostic to the specific class of an object, can compensate for this disadvantage by reducing their task to simply assigning an occupancy state to each cell in a grid, which describes a defined area around a vehicle \cite{Elfes.1989, Thrun.2005}. They usually take distance measurements, e.g. from a lidar sensor, as input. The resulting occupancy grid maps (OGMs) are for example used in the automated vehicles developed in the UNICARagil project \cite{Woopen.2018, Buchholz.2020}, from which this paper also originates.

To determine cell occupancy states from sensor data, an inverse sensor model (ISM) is required. In the past, geometric models have mostly been used \cite{Thrun.2005}. These approaches are often suitable for static and flat environments, but fail in dynamic and non-flat environments. Recent works also propose deep neural networks for this task as so-called deep ISMs \cite{Bauer.2019, Sless.2019}. As usual with supervised learning, the gathering of training data poses a challenge here. Both \cite{Bauer.2019} and \cite{Sless.2019} use cross-modal training. They make use of a lidar-based geometric ISM to generate training data for a radar-based deep ISM. This approach enables the learned model to infer occupancy information which cannot be deduced with the geometrical approach alone. Since the model is trained with data that does not constitute ground-truth data but only an estimation from the lidar-based geometric ISM, the trained models suffer from the restrictions of the lidar-based ISM.

\begin{figure}[!t]
    \centering
    \includegraphics[width=\linewidth]{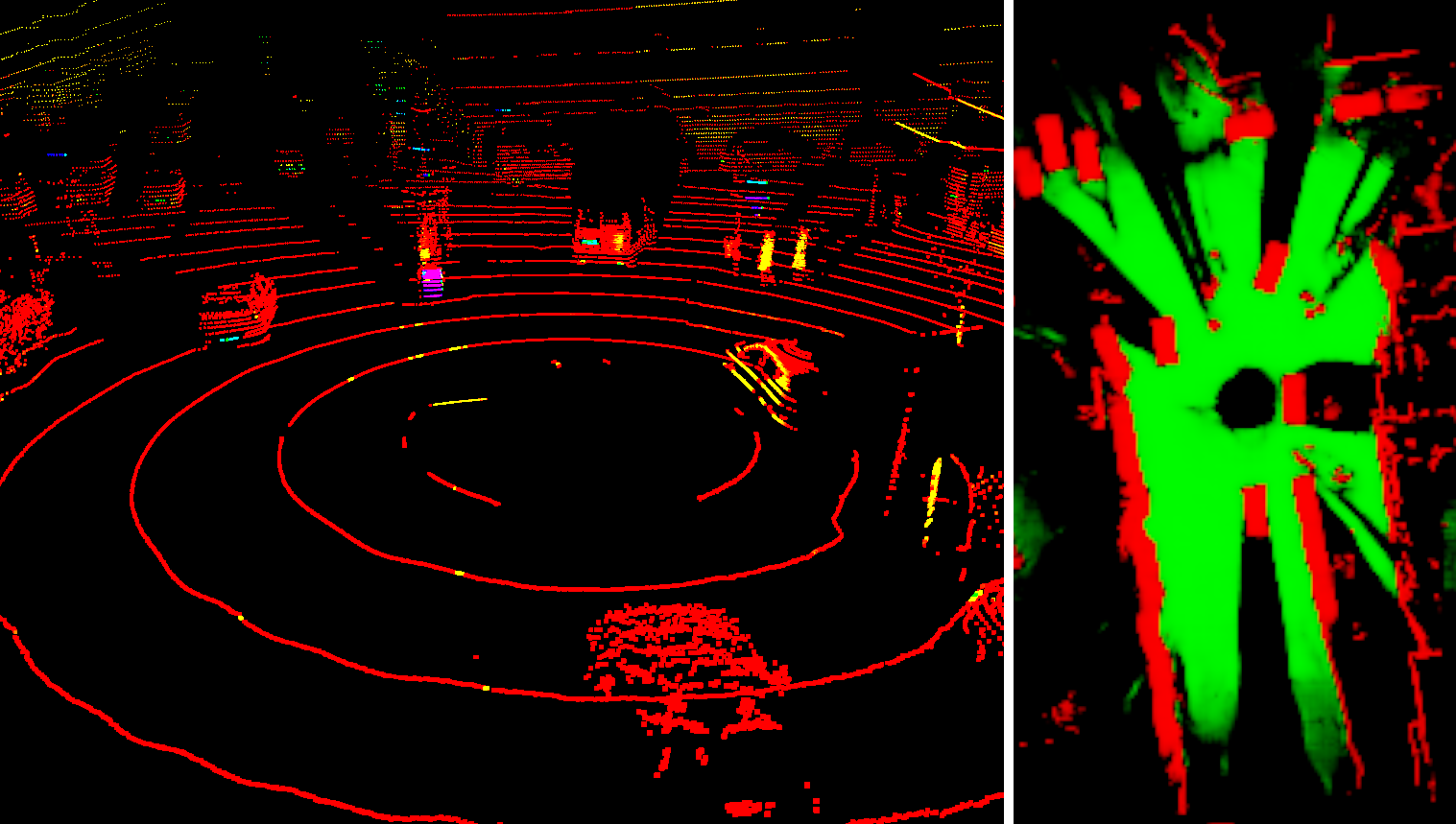}
    \caption{\label{fig_prediction}A deep learning-based inverse sensor model predicts an evidential occupancy grid map (right) from a real-world lidar point cloud (left). Grid cells with a large belief mass for the state \textit{Free} are colored green, those with a large belief mass for the state \textit{Occupied} are colored red. The bigger the belief mass, the larger the respective value in the color channel. Black represents the maximum of epistemic uncertainty.}
\end{figure}

If cross-modal training is discarded, one can make use of manually labeled data for training deep ISMs. This approach is infeasible for many though, because of the associated amount of manual labeling work. Fortunately, simulation software for automated driving is rapidly evolving. Better sensor models and the possibility to automatically generate ground-truth data make this approach viable. With more realistic synthetic data, the reality gap is closing and generalization of neural networks from synthetic to real-world data becomes possible. 

In any event, there is a need for the quantification of uncertainty in the estimates of perception algorithms. The capability of neural networks to output a measure of their confidence in a prediction is not yet reflected in many deep learning-based approaches. 

Our work takes into account all of the aforementioned challenges and contributes a framework in which they are dealt with.

\section{CONTRIBUTION}
We present an end-to-end learning framework for training deep learning-based inverse lidar sensor models using synthetic data. First, a method for generating training samples consisting of lidar point clouds as input data and evidential ground-truth OGMs as labels is presented. Second, we propose a suitable neural network architecture that extends the popular PointPillars architecture \cite{Lang.2019b} with an evidential prediction head (EPH). The EPH is capable of estimating an evidential OGM. Finally, we evaluate the performance of our approach both on synthetic data and on real-world data. We show that the trained model is able to generalize to different synthetic environments and also produces promising results on real-world data.

\section{BACKGROUND}
In the following, the concept of OGMs and an overview of current approaches for computing OGMs through inverse sensor models (ISMs) is presented. A geometric ISM will be the baseline for the evaluation of the deep ISM proposed in this work. Additionally, a brief introduction into evidence theory and subjective logic is given as this is the basis for the loss function that is used with the proposed deep neural network.

\subsection{Occupancy Grid Maps}\label{ogm}
OGMs as introduced in \cite{Elfes.1989} divide the vehicle's environment into discrete cells containing occupancy information. The occupancy state of each cell at time $k$ can be represented as a binomial random variable $ o_k \in \{ O,F \} $ ($O$: \textit{Occupied}; $F$: \textit{Free}). The probability of each cell being occupied at time $k$ can be derived from the current measurement $z_{k}$ using an inverse sensor model (ISM) $p_{z_{k}}(o_{k}|z_{k})$. Often, a binary Bayes filter is used to create an OGM from a set of distance measurements \cite{Thrun.2005}. This approach has some deficiencies. First, the binary Bayes filter relies on the Markov assumption that there is no temporal correlation of the occupancy state, and that the state does not change in time, which is not adequate for a dynamically changing environment. Additionally, by representing the occupancy as a probability value $p_k(o_k)$, it cannot be distinguished between cells which are uncertain because they cannot be observed, e.g. due to occlusions, and because of conflicting evidence as both cases are described with $p_k(o_k) \approx 0.5$. This challenge can be tackled with evidence theory as described in Section \ref{evidence_theory}.

\subsection{Inverse Sensor Models}
While a sensor model describes how the environment is represented in sensor data, an inverse sensor model (ISM) is required to reconstruct information about the environment from sensor data, as done in e.g. grid mapping.

\textbf{Geometric ISMs} process distance measurements and calculate a probability distribution for the occupancy of cells at the location of the measurement as well as for cells in between the sensor and the measurement. These models consider the sensor's inaccuracy, which is usually hand-crafted, but can also be learned \cite{Thrun.2005}. Hand-designed methods assume a ground model to separate ground from obstacles in the measurement. By filtering techniques, such as Bayesian filtering, the information gathered from each data point in the measurement can be combined into one measurement OGM. By combining the latter with a previous estimate of the OGM, a new estimate with reduced uncertainty can be found \cite{Thrun.2005}.

\textbf{Deep ISMs} are deep learning-based inverse sensor models, which take distance measurements as input for a deep neural network, which is used to predict an OGM. For this purpose, tensor-like representations for the sensor data as input and the OGM as output must be found such that they can be processed. 

In \cite{Sless.2019}, a deep radar ISM is introduced. Multiple radar measurements are combined into one bird's-eye-view image serving as input for a semantic segmentation task with classes \textit{Occupied}, \textit{Free} and \textit{Unobserved}. Training data is created from OGMs calculated from lidar measurements using a geometric ISM.

The model presented in \cite{Dequaire.2018} predicts future OGMs from lidar measurements. It is trained using unsupervised learning by creating training samples of OGMs with previous measurements. They use a naive model to create the ground-truth OGM by solely treating all reflection points in a height of $0.6$ to $1.5$ meters as obstacles. This only works in flat environments and with small pitch and roll angles of the ego-vehicle. They encode the lidar measurements into two matrices covering the vehicle's environment with binary values describing the visibility and the occupancy of each cell. The OGM is encoded as a binary matrix which does not sufficiently allow describing uncertainty.

In \cite{Bauer.2019}, radar data is transformed into a two-channel bird's-eye-view image with one channel containing static and the other containing dynamic detections. The OGM is represented as a three-channel image containing belief masses (cf. Section \ref{evidence_theory}) for the states \textit{Occupied}, \textit{Free} and \textit{Unknown}. The task is treated as image segmentation problem. In \cite{Bauer.2020}, they combine a deep radar ISM with geometrical lidar and radar ISMs to increase the perception field and to reduce the time needed to populate the occupancy grid.

The deep neural network presented in \cite{Wu.2020} predicts a bird's-eye-view image containing occupancy, object class and motion information in one shot. A sequence of lidar point clouds encoded as binary matrices which are stacked along a third dimension are used as input data. The model is trained using labeled data from the nuScenes data set \cite{Caesar.2020}.

A methodology capable of transforming segmented images from vehicle cameras to a semantic OGM is presented in~\cite{Reiher.2020}. They also use synthetic data and try to bridge the reality gap by using segmented images as an intermediate representation between real-world and synthetic sensor data.

\subsection{Evidence Theory} \label{evidence_theory}
\textbf{Evidence Theory} as introduced by Dempster and Shafer (DST) \cite{Shafer.1976} can be understood as a generalization of Bayesian probability theory \cite{Dempster.1968}. It allows the explicit consideration of epistemic uncertainty and has also been used with OGMs \cite{Nuss.2018}. Using DST, belief masses are assigned to all subsets of the frame of discernment $\Theta$. For cells in an evidential OGM, this can consist of all possible and mutually exclusive cell states \textit{Free} ($F$) and \textit{Occupied} ($O$): $\Theta = \{ F,O \}$. The power set $2^\Theta = \{ \emptyset, \{ F \}, \{ O \}, \Theta \}$ contains all possible subsets of $\Theta$ to which belief masses $m$ can be assigned.
\begin{eqnarray}
    m: 2^\Theta & \rightarrow & [0,1] \\
    m(\emptyset) & = & 0 \\
    \sum_{A \in 2^\Theta} m(A) & = & 1
\end{eqnarray}
In this example, $m(O)$ constitutes evidence for a cell being occupied and $m(F)$ for a cell being free. Additionally, a state for which no evidence is available can be addressed with $m(\Theta)$ while the empty set is no possible outcome.

\textbf{Subjective Logic} (SL) is a mathematical framework for reasoning under uncertainty. It explicitly distinguishes between epistemic opinions and aleatoric opinions. A direct bijective mapping between the belief mass distribution $m(A)$ in DST and a subjective opinion, i.e. a belief mass distribution $b_A$ and an uncertainty mass $u$ in SL is given by \cite{Jsang.2016}.
\begin{eqnarray}
    b_A & = & m(A) , \qquad A \in \Theta \\
    u & = & m(\Theta) \\
    \sum_{A \in \Theta} b_A + u & = & 1
\end{eqnarray}

Subjective opinions are equivalent to a Dirichlet probability density function (PDF)
\begin{eqnarray}
    \text{Dir}(\boldsymbol{p}, \boldsymbol{\alpha}) & = & \frac{1}{B(\boldsymbol{\alpha})} \prod_{A \in \Theta} p_A^{\alpha_A - 1} \label{eq:dirichlet_pdf}
\end{eqnarray}
 with prior probabilities $\boldsymbol{p}$ and parameters $\boldsymbol{\alpha}$
\begin{equation}
\begin{split}
    \boldsymbol{p}& = \left\{ p_A \quad | \quad A \in \Theta, \quad \sum_{A \in \Theta} p_A = 1, \quad 0 \leq p_A \leq 1 \right\} \\
    \boldsymbol{\alpha}& = \left\{ \alpha_A  \quad | \quad A \in \Theta \right\}
\end{split}
\end{equation}
and the multivariate beta function $B$ in terms of the gamma function $\Gamma$
\begin{eqnarray}
    B(\boldsymbol{\alpha}) & = & \frac{\prod_{A \in \theta} \Gamma(\alpha_A)}{\Gamma ( \sum_{A \in \Theta} \alpha_A )}.
\end{eqnarray}

Evidence for the singletons in the FOD $e_{A} \geq 1, A \in \Theta$ can be converted to parameters of a Dirichlet PDF and to a subjective opinion $(\boldsymbol{b}, u)$ with the number of classes $K = \left| \Theta \right|$ and the Dirichlet strength $S = \sum_{A \in \Theta} \alpha_{A}$:
\begin{eqnarray}
    \alpha_{A} & = & e_{A} + 1 \;, \qquad A \in \Theta \label{eq:evidence_to_alpha} \\
    b_A & = & \frac{e_A}{S} \\
    u & = & \frac{K}{S}
\end{eqnarray}

The authors of \cite{Sensoy.2018} show that a deep neural network can be trained to predict the parameters $\boldsymbol{\alpha}$ of a Dirichlet PDF to express uncertainty in a classification task on the MNIST data set \cite{LeCun.2010}. They propose three loss functions while the approach using the sum of squares led to the best results and is used in this work. The loss for one grid cell $i$ with network parameters $\boldsymbol{w}$, expected class probabilities $\boldsymbol{\hat{p}_i}$ and the true state $\boldsymbol{y_i} = \{y_{i,A} \in \{0,1\} \; | \; A \in \Theta\}$ with $y_{i,A}$ being one if state $A$ is true and zero if false or unknown is given:
\begin{equation}
    \mathcal{L}_i(\boldsymbol{w}) = \left\| \boldsymbol{y_i} - \boldsymbol{\hat{p}_i} \right\|_2^2 \text{Dir}(\boldsymbol{\hat{p}_i}, \boldsymbol{\hat{\alpha}_i}) \label{eq:dir_loss}
\end{equation}

This motivates us to create an evidential deep neural network for the task of occupancy grid mapping. We train a model to predict the parameters of a Dirichlet PDF describing the states of cells in an OGM.

\section{LEARNING FRAMEWORK}

Our learning framework processes lidar point clouds as input data and predicts evidential OGMs. In the following, the network architecture and a simulation-based method for generating and augmenting training data is presented.

\subsection{Network Architecture} \label{network_architecture}
The architecture of our deep neural network is based on the popular PointPillars architecture \cite{Lang.2019b}, which is capable of accurately detecting objects in lidar point clouds, while providing a relatively low execution time. The network is divided into three parts. First, there is the Pillar Feature Net, which encodes the reflection points from the lidar measurement into denser features. Second, there is a 2D CNN backbone, which transforms the features into a high-level representation. Last, there are detection heads estimating bounding boxes and motion states for the measured objects.

\begin{figure*}[!t]
    \centerline{
    \includegraphics[width=\textwidth]{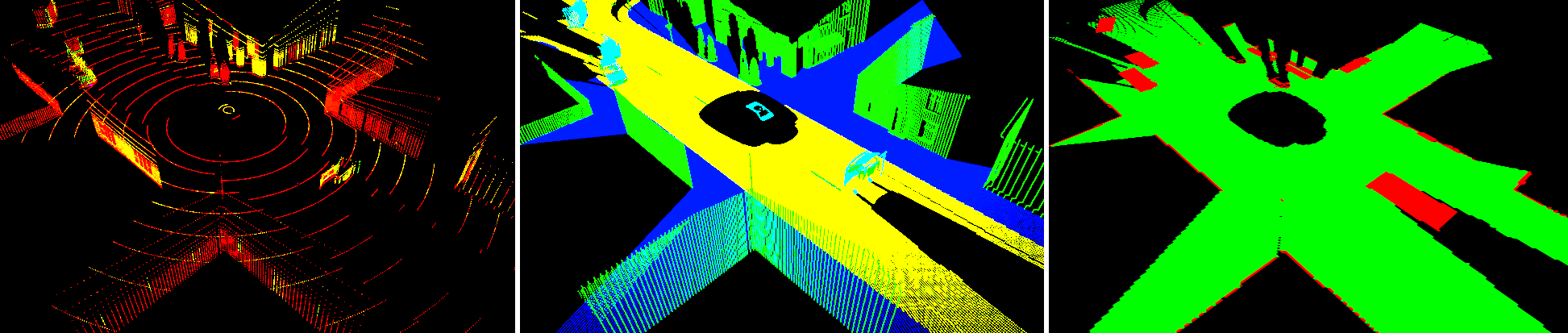}}
    \caption{The left image shows the point cloud of a simulated VLP32C lidar sensor where the point color indicates the intensity of reflection. In the middle, the corresponding high-definition (HD) point cloud with 3000 layers and points colored by the material causing the reflection is shown. This is used to create the ground-truth occupancy grid map that is shown in the right picture. Free cells are colored green, occupied cells are colored red and unknown cells are black. In addition to the information gained from the HD point cloud, all cells covered by other traffic participants are marked as occupied. A training sample consists of one point cloud as shown in the left image and a corresponding ground truth OGM as shown in the right image.}
    \label{fig_training_data}
\end{figure*}

In this work, the detection heads used in \cite{Lang.2019b} are replaced with a 2D convolutional layer with two output channels and ReLU activation. Each pixel represents a cell $i$ in the predicted OGM where the channels contain evidence $e_{i,A}$ for both singletons in the frame of discernment $\Theta = \{F,O\}$, i.e. the cell being occupied or free. The predicted evidence can be converted into estimated parameters of a Dirichlet PDF $\hat{\alpha}_{i,A}$ with Equation (\ref{eq:evidence_to_alpha}). The expected probability values of a cell being occupied or free can be derived from the Dirichlet PDF as $\hat{p}_{i,A} = \hat{\alpha}_{i,A} / S_i$. Thus, following the simplification presented in \cite{Sensoy.2018}, the loss function from Equation (\ref{eq:dir_loss}) in our application reduces to
\begin{equation}
\begin{split}
    \mathcal{L}_i(w)& = \mathbb{E} \left[ y_{i,F}^2 - 2 y_{i,F} p_{i,F} + p_{i,F}^2 \right] \\
                    & \quad + \mathbb{E} \left[ y_{i,O}^2 - 2 y_{i,O}p_{i,O} + p_{i,O}^2 \right] \\
                    & = (y_{i,F} - \hat{p}_{i,F})^2 + \frac{\hat{p}_{i,F} (1 - \hat{p}_{i,F})}{S_i + 1} \\
                    & \quad + (y_{i,O} - \hat{p}_{i,O})^2 + \frac{\hat{p}_{i,O} (1 - \hat{p}_{i,O})}{S_i + 1}.
\end{split}
\end{equation}

We extend the loss function with a Kullback-Leibler divergence term with an impact factor $\lambda_t = \min(1.0, t/10)$ that increases from zero to one with the epoch number $t$ as proposed in \cite{Sensoy.2018}. This regularization term penalizes a divergence of the Dirichlet parameters resulting from conflicting evidence $\boldsymbol{\Tilde{\alpha}_i} = \boldsymbol{y_i}+(1-\boldsymbol{y}) \odot \boldsymbol{\alpha_i}$ from a distribution resulting from no evidence, i.e. $\boldsymbol{\alpha_i} = \boldsymbol{1}$. Hence, it encourages the network to reduce conflicting evidence in its output. This leads to the total loss function for all $N$ cells of the OGM in one sample.
\begin{equation}
    \mathcal{L}(w) = \sum_{i=1}^N \mathcal{L}_i(w) + \lambda_t \text{KL}\left[ \text{Dir}(\boldsymbol{p_i}|\boldsymbol{\Tilde{\alpha}_i}) || \text{Dir}(\boldsymbol{p_i} | \boldsymbol{1}) \right]
\end{equation}
with the Kullback-Leibler divergence in terms of the gamma function $\Gamma$ and the digamma function $\psi$.
\begin{equation}
\begin{split}
    &\text{KL}\left[ \text{Dir}(\boldsymbol{p_i}|\boldsymbol{\Tilde{\alpha}_i}) || \text{Dir}(\boldsymbol{p_i} | \boldsymbol{1}) \right] = \\
    & \quad \text{log}\left( \frac{\Gamma(\Tilde{\alpha}_{iF} + \Tilde{\alpha}_{iO})}
                          {\Gamma(2) \Gamma(\Tilde{\alpha}_{iF}) \Gamma(\Tilde{\alpha}_{iO})}
              \right) \\
    & \qquad + (\Tilde{\alpha}_{iF} - 1) \left[ \psi(\Tilde{\alpha}_{iF}) - \psi\left(\Tilde{\alpha}_{iF} + \Tilde{\alpha}_{iO} \right) \right] \\
    & \qquad + (\Tilde{\alpha}_{iO} - 1) \left[ \psi(\Tilde{\alpha}_{iO}) - \psi\left(\Tilde{\alpha}_{iF} + \Tilde{\alpha}_{iO}\right) \right]
\end{split}
\end{equation}

This loss function will be used to train the model in the experiments such that it approximates the evidence masses in the grid cells while loss for occupied cells is weighted $100$ times compared to the other cells to compensate for their under-representation.

\subsection{Training Data}\label{training_data}
To create training data from the simulation, we use a method that we have already presented in \cite{Lampe.2020}. The simulation environment \cite{vonNeumannCosel.2009} provides a lidar plugin that supports advanced ray tracing and physically-based rendering, i.e. it considers the physical properties of materials to generate more realistic point clouds. We have modeled the sensor setup of one of our research vehicles, which has a Velodyne VLP32C lidar sensor mounted in the middle of its roof. In addition to this 32-layer lidar sensor, another lidar sensor with 3000 layers, which will be called high-definition (HD) lidar from now on, was added to the simulated vehicle. As the lidar plugin provides information about the material type that caused a reflection, it is possible to derive a dense ground-truth OGM from the simulated HD lidar data.  The HD lidar covers the same field of view and is exposed to the same occlusions as the real sensor to ensure that the OGMs in the training data will only contain information which can theoretically be derived from the input point cloud. In addition, we augment the ground-truth OGMs based on a ground-truth object list that is also provided by the simulation and contains information about the position and shape of other traffic participants. As we expect the deep ISM to learn to recognize the shape of e.g. cars and trucks, we mark all cells covered by them as occupied in the ground-truth OGM if a minimum of 50 reflections on the vehicle is present in the input point cloud. Figure \ref{fig_training_data} shows that cells with reflections on obstacles such as cars and trees are marked as occupied in the ground-truth OGM, whereas the ground is marked as free. We decided to mark also sidewalks as free space as precise information on the lane geometry can be obtained from a static map. In the event of a serious obstruction of road traffic, the sidewalk could also be an option to pass by.

\subsection{Augmentation}
During the training, we augment the training data by applying a random rotation to the input point cloud and the label OGM. Both are rotated around a vertical axis at the origin of the lidar sensor. This turned out to be important, as without augmentation, cells on the far left and right side of the OGM are almost always occupied, which makes it difficult for the deep ISM to classify those cells correctly.

\section{EXPERIMENTAL SETUP AND RESEARCH QUESTION}
As explained in Section \ref{training_data}, training data for the deep learning-based inverse lidar model is created using a simulation environment. We have selected a model of an urban area and created ten scenarios, each containing random variations of the dynamic environment. In half of the scenarios, the ego-vehicle takes a clockwise route, in the other half, it takes a counter-clockwise route. The randomly generated pulk traffic contains cars, trucks and motorcycles. The type of each of these objects is also chosen randomly from a larger catalogue of possibilities. Additionally, pedestrians are placed at various locations on the sidewalks. Parked vehicles are randomly put on parking lanes. With each scenario, $1.000$ training samples were created at a sampling rate of one per second. In total, $10.000$ samples were generated for training. For the validation data set, another $1000$ samples were created in the same static, but a different dynamic environment. Another test dataset consists of $100$ samples from a different scenario in another part of the simulated urban area and will be used for evaluation in Section \ref{eval_syn}.

The neural network is trained using the Adam optimizer. The Pillar Feature Net \cite{Lang.2019b} creates $10.000$ pillars with a maximum of $100$ reflection points per pillar. The intensity of the reflection points in the simulated lidar point cloud is normalized such that a distribution similar to one observed in real-world lidar point clouds is achieved.  The output grid maps have a length of $81.92$ and a width of $56.32$ meters. A cell's side length is $16$ centimeters, resulting in a $512$ by $352$ cell grid map. The sensor origin is at the center of the grid map. The map dimensions and cell size correspond to the detection area and step size of the Pillar Feature Net. After training for $100$ epochs with a batch size of $5$, a minimum loss of $0.104$ and a Kullback-Leibler divergence of $0.357$ on the validation data was achieved.

In the following, we want to answer the following research questions: How well does a deep convolutional neural network that is based on our presented methodology perform when predicting dense OGMs from lidar measurements? In particular, we want to analyze whether it is capable of capturing the epistemic uncertainty for cells where no reflection point is located. Then, how well does the network perform when presented with real-world sensor data?

\section{RESULTS AND DISCUSSION}
First, we evaluate the performance of the trained model on a synthetic test dataset to analyze how well the trained ISM predicts OGMs in a scenario that was not contained in the training data.

Afterwards, we test our model on real-world sensor data that was recorded with one of our research vehicles.

\begin{figure}[!h]
    \center
    \includegraphics[width=\linewidth]{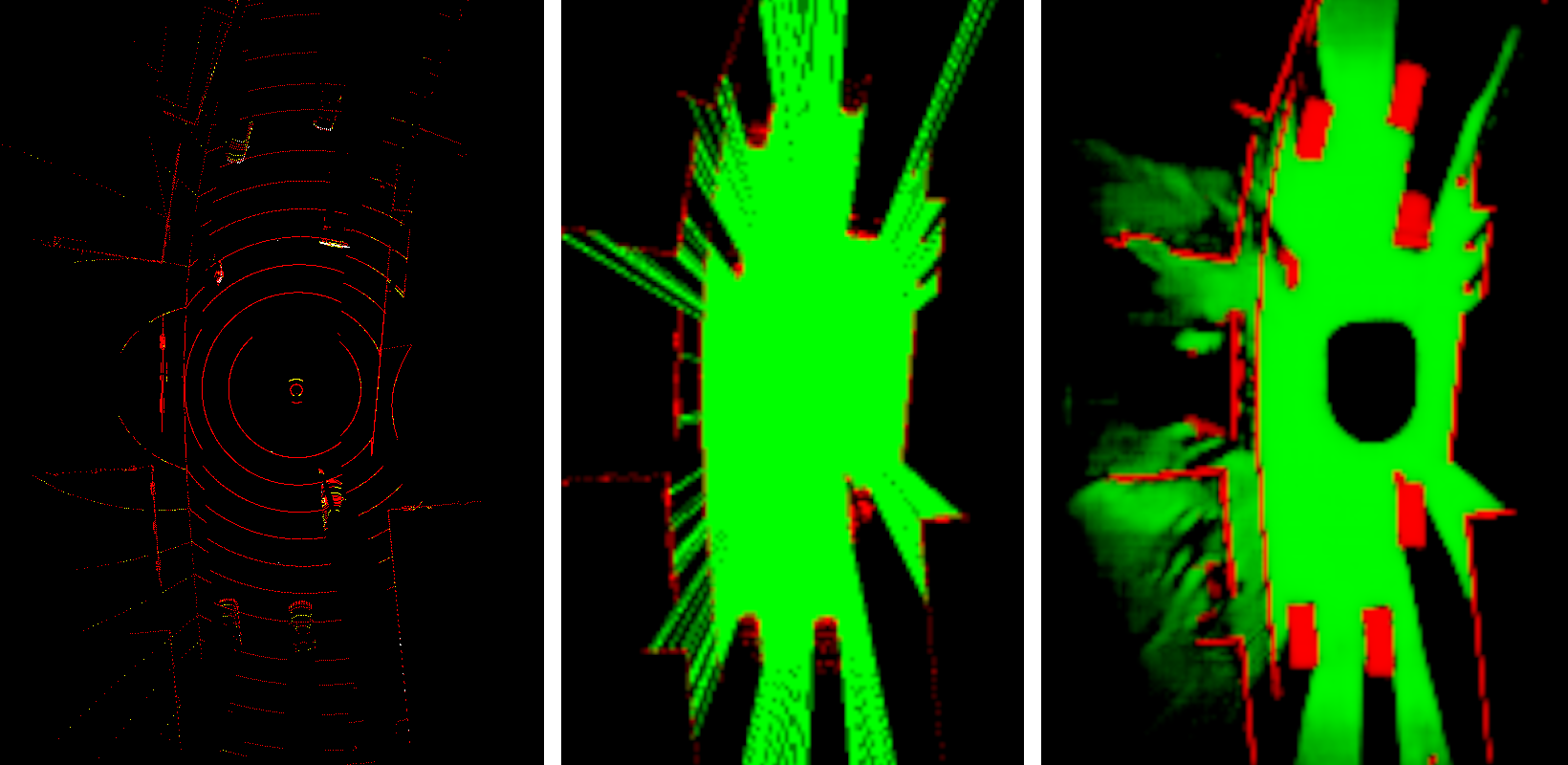}
    \caption{\label{fig_naive}The geometric ISM (middle) is only able to determine occupied cells containing reflection points of the measurement (left). The deep learning-based ISM (right) has learned to derive more information, e.g. the whole space occupied by vehicles.}
\end{figure}

\subsection{Evaluation on Synthetic Data} \label{eval_syn}
\input{evaluation_plot}

We compare the OGMs that are created using our proposed deep ISM to OGMs created using a geometric ISM where all cells containing reflection points in a height of $0.5$ to $2.0$ meters above ground are marked as occupied while all cells between the sensor origin and these cells are marked as free. Figure \ref{fig_evaluation_masses} shows the mean belief masses in both OGMs during the test scenario. It is apparent and confirmed by Figure \ref{fig_naive} that the geometric ISM creates a considerably higher proportion of cells with an unknown state, hence, less cells are classified as free or occupied. Figure \ref{fig_evaluation_kld} shows the Kullback-Leibler divergence of both Dirichlet PDFs, the one predicted from the deep ISM $\text{Dir}(\boldsymbol{p}|\boldsymbol{\hat{\alpha}})$ and the one from the geometric ISM $\text{Dir}(\boldsymbol{p}|\boldsymbol{\hat{\alpha}_G})$, from the true PDF $\text{Dir}(\boldsymbol{p}|\boldsymbol{\alpha})$. It is evident that the deep ISM estimates the cell states better than the geometric ISM at any time.

\subsection{Evaluation on Real-World Data}
\begin{figure}[!ht]
    \center
    \includegraphics[width=0.9\linewidth]{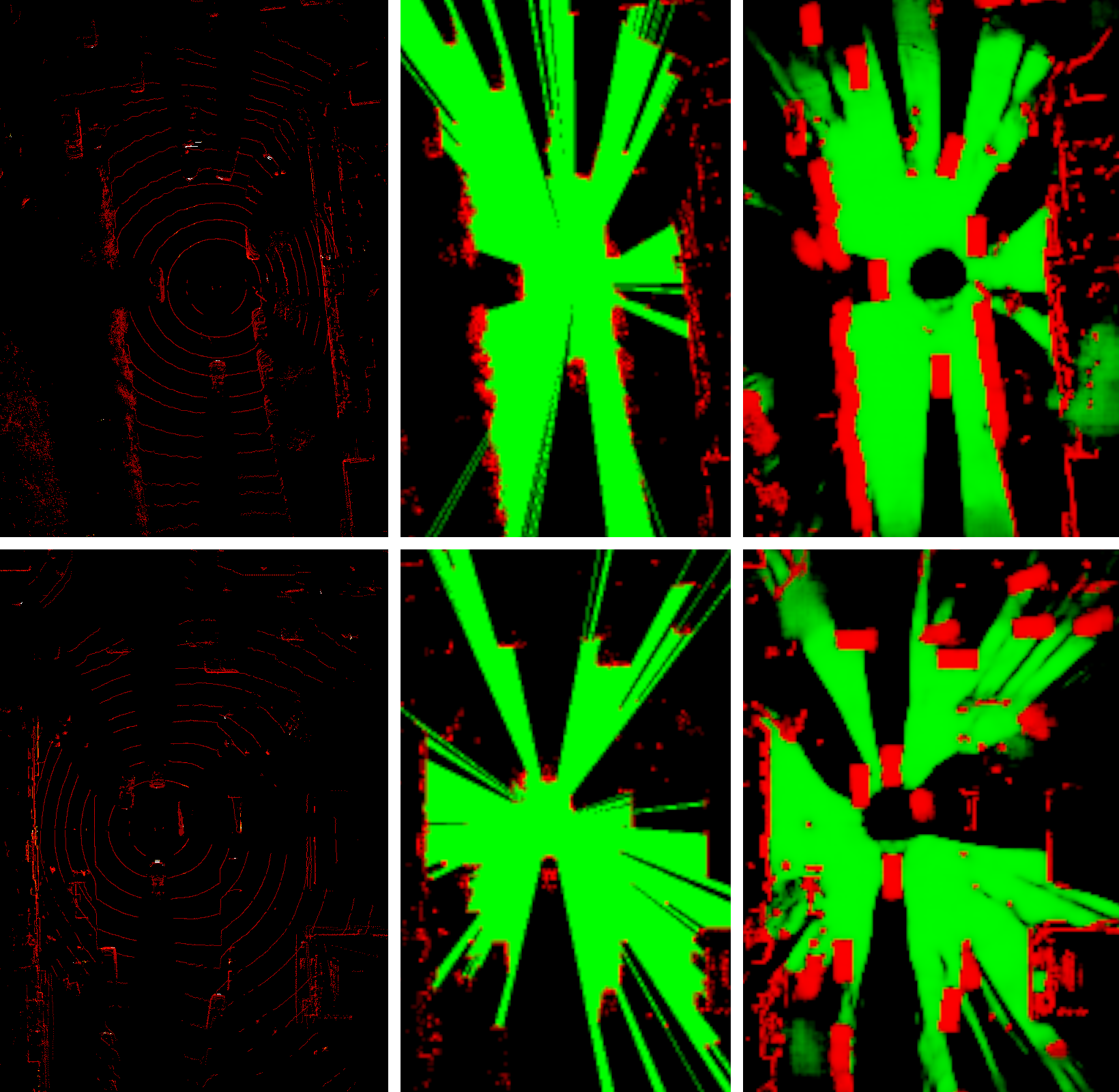}
    \caption{\label{fig_real}Two representative samples showing the predicted OGMs (right) using the deep ISM trained with synthetic data on real-world measurements (left) compared to a geometric ISM (middle).}
\end{figure}
Finally, we want to analyze whether the deep ISM trained with synthetic data can also be successfully applied to real-world data. Thus, we test the model on real lidar measurements from one of our research vehicles that has a similar shape and sensor setup as the simulated vehicle. Figure \ref{fig_real} shows two representative samples of the model's performance on a dataset recorded in an urban environment. It is apparent that the performance cannot keep up with the evaluation on synthetic data. Nevertheless, it creates promising results in an environment that is considerably different from the synthetic environment. In particular, the occupancy states of cells that do not contain a reflection point are mostly rated correctly. A higher real-world performance is expected when the simulation scenarios would take place in a model of the actual static environment of the real-world scenarios. More samples and a higher diversity of the training data are relatively easy to achieve in the simulation and are also expected to further increase performance.

\section{CONCLUSION}
We presented a new methodology to train neural networks for the task of occupancy grid mapping using lidar point clouds. The trained network performs considerably better than a classical approach when presented with synthetic data. It also shows advantages when presented with real-world data, even though many options to increase the network's generalization capabilities remain. In contrast to many other deep learning-based approaches, our network is trained such that it can give an estimate of its first- and second-order uncertainty. Our methodology is especially promising because it does not rely on manually labeled data. Future research will investigate methods that further increase the network's performance on real-world data.
\vfill\null

\addtolength{\textheight}{-4cm}   






\bibliographystyle{IEEEtran}
\bibliography{literature}

\end{document}

%% file: evaluation_plot.tex
\begin{figure}[!ht]
    \centering
	\begin{tikzpicture}
	\begin{axis}[
    smooth, enlarge x limits=false, xmin=0, xmax=100, ymin=0.0, ymax=1.0, 
	xlabel={time in seconds},
	y=4cm,
	x label style={at={(axis description cs:0.5,0.02)},anchor=north},
	legend entries={$\overline{m}(\Theta)$,  $\overline{m}_G(\Theta)$, $\overline{m}(F)$, $\overline{m}_G(F)$, $\overline{m}(O)$, $\overline{m}_G(O)$},
	legend style={at={(0.5,0.66)},anchor=north},
	legend cell align={left},
	legend columns=2,]
	\addplot+[mark=none, blue!100!white, solid, style={very thick}] coordinates 
	{(0.0, 0.8101961016654968) (1.0, 0.7611374258995056) (2.0, 0.8288246393203735) (3.0, 0.7984712719917297) (4.0, 0.7290783524513245) (5.0, 0.7623830437660217) (6.0, 0.7022638916969299) (7.0, 0.7376742959022522) (8.0, 0.790397047996521) (9.0, 0.7963049411773682) (10.0, 0.8067687749862671) (11.0, 0.7560259103775024) (12.0, 0.807847797870636) (13.0, 0.7842805981636047) (14.0, 0.8170869946479797) (15.0, 0.7717746496200562) (16.0, 0.7879853844642639) (17.0, 0.7086304426193237) (18.0, 0.7781572937965393) (19.0, 0.8268581628799438) (20.0, 0.7587867975234985) (21.0, 0.6876682639122009) (22.0, 0.7454917430877686) (23.0, 0.806511640548706) (24.0, 0.8188116550445557) (25.0, 0.7439176440238953) (26.0, 0.7624749541282654) (27.0, 0.8106843829154968) (28.0, 0.7664339542388916) (29.0, 0.7785394787788391) (30.0, 0.7508621215820312) (31.0, 0.7091732025146484) (32.0, 0.800663411617279) (33.0, 0.6985984444618225) (34.0, 0.727010190486908) (35.0, 0.7193813920021057) (36.0, 0.803871750831604) (37.0, 0.7741666436195374) (38.0, 0.7684009075164795) (39.0, 0.8195379376411438) (40.0, 0.8153062462806702) (41.0, 0.7674248218536377) (42.0, 0.7925569415092468) (43.0, 0.7716302275657654) (44.0, 0.793809175491333) (45.0, 0.7904188632965088) (46.0, 0.7017098069190979) (47.0, 0.8111526966094971) (48.0, 0.81640625) (49.0, 0.7156266570091248) (50.0, 0.6998621821403503) (51.0, 0.7577440142631531) (52.0, 0.7351216673851013) (53.0, 0.7890720367431641) (54.0, 0.8142036199569702) (55.0, 0.7705468535423279) (56.0, 0.8107948303222656) (57.0, 0.7692832350730896) (58.0, 0.7156857252120972) (59.0, 0.7101000547409058) (60.0, 0.7733093500137329) (61.0, 0.7667922973632812) (62.0, 0.8311662077903748) (63.0, 0.7296394109725952) (64.0, 0.7050208449363708) (65.0, 0.7913945913314819) (66.0, 0.7854245901107788) (67.0, 0.7693607211112976) (68.0, 0.7644088864326477) (69.0, 0.7463333010673523) (70.0, 0.7508264183998108) (71.0, 0.7256917953491211) (72.0, 0.7656794190406799) (73.0, 0.688125729560852) (74.0, 0.8386011719703674) (75.0, 0.8120656609535217) (76.0, 0.8156614303588867) (77.0, 0.7273836731910706) (78.0, 0.6856091618537903) (79.0, 0.8073593378067017) (80.0, 0.7595465183258057) (81.0, 0.7928013205528259) (82.0, 0.747897207736969) (83.0, 0.8080847263336182) (84.0, 0.7350353598594666) (85.0, 0.7988007068634033) (86.0, 0.7870067358016968) (87.0, 0.7749179601669312) (88.0, 0.7423936724662781) (89.0, 0.7343829870223999) (90.0, 0.7512251138687134) (91.0, 0.8145925402641296) (92.0, 0.7575664520263672) (93.0, 0.6920843124389648) (94.0, 0.7633097171783447) (95.0, 0.7701593041419983) (96.0, 0.759361982345581) (97.0, 0.823434591293335) (98.0, 0.8146881461143494) (99.0, 0.7076106667518616)};
	
	\addplot+[mark=none, blue!100!white, dotted, style={very thick}] coordinates 
	{(0.0, 0.9211452603340149) (1.0, 0.9194514751434326) (2.0, 0.9333255290985107) (3.0, 0.9327049255371094) (4.0, 0.9278560280799866) (5.0, 0.915143609046936) (6.0, 0.9114574790000916) (7.0, 0.9067350029945374) (8.0, 0.9239000678062439) (9.0, 0.9212730526924133) (10.0, 0.9263980388641357) (11.0, 0.9108526110649109) (12.0, 0.9278025031089783) (13.0, 0.9117650389671326) (14.0, 0.9275534152984619) (15.0, 0.9116098880767822) (16.0, 0.9198988080024719) (17.0, 0.9076361060142517) (18.0, 0.9142155051231384) (19.0, 0.9330737590789795) (20.0, 0.9128171801567078) (21.0, 0.9119419455528259) (22.0, 0.9147121906280518) (23.0, 0.9257287383079529) (24.0, 0.9344196319580078) (25.0, 0.9024944305419922) (26.0, 0.9027683734893799) (27.0, 0.9267410635948181) (28.0, 0.9077991843223572) (29.0, 0.9160570502281189) (30.0, 0.9279151558876038) (31.0, 0.9226806163787842) (32.0, 0.9203501343727112) (33.0, 0.9200263619422913) (34.0, 0.9140238165855408) (35.0, 0.9136131405830383) (36.0, 0.9271951913833618) (37.0, 0.9137550592422485) (38.0, 0.912947416305542) (39.0, 0.9292011260986328) (40.0, 0.9258003234863281) (41.0, 0.9145860075950623) (42.0, 0.9236156344413757) (43.0, 0.9200926423072815) (44.0, 0.9159166216850281) (45.0, 0.9305927157402039) (46.0, 0.899945855140686) (47.0, 0.9342166781425476) (48.0, 0.9264400601387024) (49.0, 0.9273387789726257) (50.0, 0.8991018533706665) (51.0, 0.9073730111122131) (52.0, 0.9091801047325134) (53.0, 0.9205766320228577) (54.0, 0.9274335503578186) (55.0, 0.9110347628593445) (56.0, 0.9245933294296265) (57.0, 0.9063865542411804) (58.0, 0.9200170636177063) (59.0, 0.9033752083778381) (60.0, 0.9144534468650818) (61.0, 0.9049133062362671) (62.0, 0.9371519088745117) (63.0, 0.8971245288848877) (64.0, 0.9250515103340149) (65.0, 0.9326735138893127) (66.0, 0.9198379516601562) (67.0, 0.9196222424507141) (68.0, 0.9161956310272217) (69.0, 0.9039924144744873) (70.0, 0.903499186038971) (71.0, 0.9164844751358032) (72.0, 0.9233798980712891) (73.0, 0.9102535247802734) (74.0, 0.9371572732925415) (75.0, 0.9248085021972656) (76.0, 0.9292904138565063) (77.0, 0.9104904532432556) (78.0, 0.9098390340805054) (79.0, 0.9239823818206787) (80.0, 0.9119108319282532) (81.0, 0.9199779629707336) (82.0, 0.9108307361602783) (83.0, 0.9249420166015625) (84.0, 0.9167813658714294) (85.0, 0.9216121435165405) (86.0, 0.9320690035820007) (87.0, 0.9136756658554077) (88.0, 0.9186128973960876) (89.0, 0.906117856502533) (90.0, 0.9213428497314453) (91.0, 0.9240223169326782) (92.0, 0.9116337299346924) (93.0, 0.9170869588851929) (94.0, 0.9119800329208374) (95.0, 0.904701292514801) (96.0, 0.912532389163971) (97.0, 0.9311330914497375) (98.0, 0.9298552870750427) (99.0, 0.9016608595848083)};
	
	\addplot+[mark=none, green!100!white!60!black, solid, style={very thick}] coordinates 
	{(0.0, 0.1641995906829834) (1.0, 0.20676808059215546) (2.0, 0.14566200971603394) (3.0, 0.1674441695213318) (4.0, 0.24798302352428436) (5.0, 0.213027223944664) (6.0, 0.27395230531692505) (7.0, 0.23864375054836273) (8.0, 0.18202361464500427) (9.0, 0.1766224354505539) (10.0, 0.16056662797927856) (11.0, 0.21656584739685059) (12.0, 0.16872155666351318) (13.0, 0.18461395800113678) (14.0, 0.15062081813812256) (15.0, 0.19898878037929535) (16.0, 0.18572811782360077) (17.0, 0.26385489106178284) (18.0, 0.19051103293895721) (19.0, 0.13984502851963043) (20.0, 0.22071309387683868) (21.0, 0.28658822178840637) (22.0, 0.22683605551719666) (23.0, 0.1619383841753006) (24.0, 0.15105904638767242) (25.0, 0.22310835123062134) (26.0, 0.21012061834335327) (27.0, 0.1661049872636795) (28.0, 0.20330758392810822) (29.0, 0.1933874636888504) (30.0, 0.21334046125411987) (31.0, 0.266019731760025) (32.0, 0.17578977346420288) (33.0, 0.2738167643547058) (34.0, 0.2521103024482727) (35.0, 0.25098317861557007) (36.0, 0.16728971898555756) (37.0, 0.18839006125926971) (38.0, 0.21187062561511993) (39.0, 0.15735368430614471) (40.0, 0.15599536895751953) (41.0, 0.2021036297082901) (42.0, 0.1721428483724594) (43.0, 0.20115524530410767) (44.0, 0.1735057830810547) (45.0, 0.174720898270607) (46.0, 0.28018975257873535) (47.0, 0.1556445062160492) (48.0, 0.16301582753658295) (49.0, 0.26104775071144104) (50.0, 0.2723137140274048) (51.0, 0.2080390304327011) (52.0, 0.24723564088344574) (53.0, 0.18131019175052643) (54.0, 0.16111530363559723) (55.0, 0.19827227294445038) (56.0, 0.16400636732578278) (57.0, 0.20417222380638123) (58.0, 0.262102872133255) (59.0, 0.2611963450908661) (60.0, 0.19332660734653473) (61.0, 0.20072263479232788) (62.0, 0.13843272626399994) (63.0, 0.24142593145370483) (64.0, 0.270769327878952) (65.0, 0.175337553024292) (66.0, 0.1918160766363144) (67.0, 0.20757867395877838) (68.0, 0.21154499053955078) (69.0, 0.22774596512317657) (70.0, 0.22387383878231049) (71.0, 0.2559562027454376) (72.0, 0.20303145051002502) (73.0, 0.28695443272590637) (74.0, 0.13572493195533752) (75.0, 0.1663990020751953) (76.0, 0.15222829580307007) (77.0, 0.2540343999862671) (78.0, 0.28802958130836487) (79.0, 0.1715211272239685) (80.0, 0.21077562868595123) (81.0, 0.1735682338476181) (82.0, 0.22441668808460236) (83.0, 0.17025429010391235) (84.0, 0.23750348389148712) (85.0, 0.18106748163700104) (86.0, 0.17819112539291382) (87.0, 0.19982756674289703) (88.0, 0.22302788496017456) (89.0, 0.24077977240085602) (90.0, 0.21623530983924866) (91.0, 0.1629798859357834) (92.0, 0.22461646795272827) (93.0, 0.2828115224838257) (94.0, 0.20478886365890503) (95.0, 0.20122994482517242) (96.0, 0.20603038370609283) (97.0, 0.14527378976345062) (98.0, 0.14991217851638794) (99.0, 0.2514282763004303)};
	
	\addplot+[mark=none, green!100!white!60!black, dotted, style={very thick}] coordinates 
	{(0.0, 0.07644245028495789) (1.0, 0.07787717878818512) (2.0, 0.06421459466218948) (3.0, 0.06433723121881485) (4.0, 0.06957024335861206) (5.0, 0.08216898888349533) (6.0, 0.08595090359449387) (7.0, 0.0907655879855156) (8.0, 0.07348418235778809) (9.0, 0.07618176937103271) (10.0, 0.07077247649431229) (11.0, 0.086579829454422) (12.0, 0.07006294280290604) (13.0, 0.08521821349859238) (14.0, 0.06935190409421921) (15.0, 0.08571378886699677) (16.0, 0.07749228924512863) (17.0, 0.0899781808257103) (18.0, 0.08325731009244919) (19.0, 0.06374465674161911) (20.0, 0.08484725654125214) (21.0, 0.08530854433774948) (22.0, 0.08282895386219025) (23.0, 0.07143668830394745) (24.0, 0.06296373158693314) (25.0, 0.09457559138536453) (26.0, 0.09459180384874344) (27.0, 0.07085471600294113) (28.0, 0.0894879698753357) (29.0, 0.08156796544790268) (30.0, 0.06901279091835022) (31.0, 0.07487393915653229) (32.0, 0.07725333422422409) (33.0, 0.07759339362382889) (34.0, 0.08365510404109955) (35.0, 0.08237159997224808) (36.0, 0.07023880630731583) (37.0, 0.082997627556324) (38.0, 0.08515714854001999) (39.0, 0.06841360777616501) (40.0, 0.07156634330749512) (41.0, 0.08265725523233414) (42.0, 0.07305216044187546) (43.0, 0.07721853256225586) (44.0, 0.0810088962316513) (45.0, 0.06663797050714493) (46.0, 0.09829762578010559) (47.0, 0.06299705058336258) (48.0, 0.07164072245359421) (49.0, 0.0704646185040474) (50.0, 0.09872507303953171) (51.0, 0.08993091434240341) (52.0, 0.08909089118242264) (53.0, 0.07660812884569168) (54.0, 0.0704265609383583) (55.0, 0.08608148247003555) (56.0, 0.07315908372402191) (57.0, 0.0911109447479248) (58.0, 0.07792004942893982) (59.0, 0.09394056349992752) (60.0, 0.08244399726390839) (61.0, 0.09209876507520676) (62.0, 0.05975095182657242) (63.0, 0.10019621253013611) (64.0, 0.07246748358011246) (65.0, 0.06429720669984818) (66.0, 0.07810723781585693) (67.0, 0.07841046154499054) (68.0, 0.08176416158676147) (69.0, 0.09347378462553024) (70.0, 0.09422152489423752) (71.0, 0.08161843568086624) (72.0, 0.07337477803230286) (73.0, 0.08755571395158768) (74.0, 0.06038909777998924) (75.0, 0.07315453886985779) (76.0, 0.06759922951459885) (77.0, 0.08761665970087051) (78.0, 0.08750884234905243) (79.0, 0.0741160437464714) (80.0, 0.08544043451547623) (81.0, 0.07676798850297928) (82.0, 0.08674363046884537) (83.0, 0.0728556215763092) (84.0, 0.0808388963341713) (85.0, 0.07649143785238266) (86.0, 0.0649486854672432) (87.0, 0.0836426243185997) (88.0, 0.07834425568580627) (89.0, 0.09159315377473831) (90.0, 0.07570996135473251) (91.0, 0.07368577271699905) (92.0, 0.08651753515005112) (93.0, 0.08070068806409836) (94.0, 0.08549930900335312) (95.0, 0.09266497939825058) (96.0, 0.08444396406412125) (97.0, 0.06592833250761032) (98.0, 0.06691364198923111) (99.0, 0.09397253394126892)};
	
	\addplot+[mark=none, red!100!white!60!black, solid, style={very thick}] coordinates 
	{(0.0, 0.025604361668229103) (1.0, 0.032094478607177734) (2.0, 0.02551337331533432) (3.0, 0.03408447653055191) (4.0, 0.02293861098587513) (5.0, 0.024589765816926956) (6.0, 0.023783832788467407) (7.0, 0.023681992664933205) (8.0, 0.02757936716079712) (9.0, 0.02707260102033615) (10.0, 0.03266466036438942) (11.0, 0.027408242225646973) (12.0, 0.023430651053786278) (13.0, 0.031105494126677513) (14.0, 0.03229222074151039) (15.0, 0.02923661656677723) (16.0, 0.02628655917942524) (17.0, 0.027514664456248283) (18.0, 0.031331680715084076) (19.0, 0.03329680487513542) (20.0, 0.020500101149082184) (21.0, 0.025743460282683372) (22.0, 0.027672179043293) (23.0, 0.03154999390244484) (24.0, 0.030129259452223778) (25.0, 0.03297397866845131) (26.0, 0.02740446850657463) (27.0, 0.023210642859339714) (28.0, 0.03025841899216175) (29.0, 0.028073089197278023) (30.0, 0.035797473043203354) (31.0, 0.024807089939713478) (32.0, 0.02354680933058262) (33.0, 0.027584822848439217) (34.0, 0.020879486575722694) (35.0, 0.02963545173406601) (36.0, 0.02883853204548359) (37.0, 0.037443291395902634) (38.0, 0.01972850412130356) (39.0, 0.023108357563614845) (40.0, 0.028698433190584183) (41.0, 0.030471540987491608) (42.0, 0.035300251096487045) (43.0, 0.02721460349857807) (44.0, 0.03268502652645111) (45.0, 0.03486026078462601) (46.0, 0.018100405111908913) (47.0, 0.033202800899744034) (48.0, 0.020577900111675262) (49.0, 0.023325635120272636) (50.0, 0.027824146673083305) (51.0, 0.034216973930597305) (52.0, 0.01764274761080742) (53.0, 0.029617737978696823) (54.0, 0.02468099445104599) (55.0, 0.03118087165057659) (56.0, 0.025198854506015778) (57.0, 0.026544513180851936) (58.0, 0.022211361676454544) (59.0, 0.028703559190034866) (60.0, 0.033364009112119675) (61.0, 0.032485101372003555) (62.0, 0.030401062220335007) (63.0, 0.028934674337506294) (64.0, 0.024209778755903244) (65.0, 0.03326786682009697) (66.0, 0.022759370505809784) (67.0, 0.023060591891407967) (68.0, 0.02404610440135002) (69.0, 0.025920726358890533) (70.0, 0.0252997986972332) (71.0, 0.01835199072957039) (72.0, 0.03128904476761818) (73.0, 0.024919845163822174) (74.0, 0.02567383460700512) (75.0, 0.021535396575927734) (76.0, 0.0321102999150753) (77.0, 0.01858193799853325) (78.0, 0.026361210271716118) (79.0, 0.021119533106684685) (80.0, 0.02967783994972706) (81.0, 0.03363041952252388) (82.0, 0.02768607623875141) (83.0, 0.02166096679866314) (84.0, 0.027461154386401176) (85.0, 0.020131828263401985) (86.0, 0.03480212390422821) (87.0, 0.025254549458622932) (88.0, 0.034578464925289154) (89.0, 0.024837207049131393) (90.0, 0.032539594918489456) (91.0, 0.02242756076157093) (92.0, 0.017817014828324318) (93.0, 0.02510417066514492) (94.0, 0.03190141171216965) (95.0, 0.02861076593399048) (96.0, 0.03460759297013283) (97.0, 0.03129168599843979) (98.0, 0.035399600863456726) (99.0, 0.04096103087067604)};
	
	\addplot+[mark=none, red!100!white!60!black, dotted, style={very thick}] coordinates 
	{(0.0, 0.0024122418835759163) (1.0, 0.002671417547389865) (2.0, 0.0024598841555416584) (3.0, 0.0029579205438494682) (4.0, 0.0025737332180142403) (5.0, 0.002687365049496293) (6.0, 0.002591543598100543) (7.0, 0.0024994800332933664) (8.0, 0.0026157463435083628) (9.0, 0.002545206807553768) (10.0, 0.002829565666615963) (11.0, 0.0025676139630377293) (12.0, 0.0021345692221075296) (13.0, 0.003016839735209942) (14.0, 0.0030946852639317513) (15.0, 0.0026762995403259993) (16.0, 0.002608983078971505) (17.0, 0.0023857064079493284) (18.0, 0.0025272145867347717) (19.0, 0.0031816863920539618) (20.0, 0.0023356296587735415) (21.0, 0.0027495103422552347) (22.0, 0.0024588017258793116) (23.0, 0.002834656275808811) (24.0, 0.0026166231837123632) (25.0, 0.002930079586803913) (26.0, 0.0026398112531751394) (27.0, 0.0024041691794991493) (28.0, 0.002712809480726719) (29.0, 0.002375067677348852) (30.0, 0.003072091145440936) (31.0, 0.0024454728700220585) (32.0, 0.0023965141735970974) (33.0, 0.002380229765549302) (34.0, 0.002321184379979968) (35.0, 0.00401531346142292) (36.0, 0.002566027455031872) (37.0, 0.003247380955144763) (38.0, 0.0018954392289742827) (39.0, 0.002385273575782776) (40.0, 0.0026333765126764774) (41.0, 0.0027568272780627012) (42.0, 0.0033322537783533335) (43.0, 0.0026887860149145126) (44.0, 0.003074543783441186) (45.0, 0.0027693226002156734) (46.0, 0.0017565421294420958) (47.0, 0.002786348108202219) (48.0, 0.0019192370818927884) (49.0, 0.0021966174244880676) (50.0, 0.002173075219616294) (51.0, 0.002696116454899311) (52.0, 0.0017290073446929455) (53.0, 0.002815314568579197) (54.0, 0.0021399958059191704) (55.0, 0.002883798675611615) (56.0, 0.002247649012133479) (57.0, 0.0025025182403624058) (58.0, 0.002062874846160412) (59.0, 0.002684284234419465) (60.0, 0.003102612681686878) (61.0, 0.002987983636558056) (62.0, 0.003097147447988391) (63.0, 0.002679301891475916) (64.0, 0.0024809606838971376) (65.0, 0.0030292647425085306) (66.0, 0.002054833807051182) (67.0, 0.0019673688802868128) (68.0, 0.002040307968854904) (69.0, 0.0025338265113532543) (70.0, 0.0022792795207351446) (71.0, 0.0018971319077536464) (72.0, 0.00324534485116601) (73.0, 0.0021907377522438765) (74.0, 0.0024536289274692535) (75.0, 0.0020370588172227144) (76.0, 0.003110353834927082) (77.0, 0.0018928772769868374) (78.0, 0.002652161754667759) (79.0, 0.0019016076112166047) (80.0, 0.0026488315779715776) (81.0, 0.0032540932297706604) (82.0, 0.0024255658499896526) (83.0, 0.002202344126999378) (84.0, 0.002379769692197442) (85.0, 0.0018964071059599519) (86.0, 0.002982315607368946) (87.0, 0.0026817789766937494) (88.0, 0.0030428553000092506) (89.0, 0.002288932679221034) (90.0, 0.002947188215330243) (91.0, 0.0022919608745723963) (92.0, 0.0018487311899662018) (93.0, 0.002212418243288994) (94.0, 0.0025206319987773895) (95.0, 0.0026337015442550182) (96.0, 0.0030236816965043545) (97.0, 0.0029386328533291817) (98.0, 0.0032310648821294308) (99.0, 0.004366710316389799)};
	
    \end{axis}
	\end{tikzpicture}
	\caption{\label{fig_evaluation_masses}Mean belief masses per OGM during the test scenario. $\overline{m}(F)$ is the mean mass for a free, $\overline{m}(O)$ for an occupied and $\overline{m}(\Theta)$ for an unknown cell state. The dashed lines represent results based on a geometric ISM, whereas the continuous lines show results from our deep ISM.}
\end{figure}

\begin{figure}[!ht]
    \centering
	\begin{tikzpicture}
	\begin{axis}[
    smooth, enlarge x limits=false, xmin=0, xmax=100, ymin=0.0, ymax=2.5,
    y=1.5cm,
	xlabel={time in seconds},
	x label style={at={(axis description cs:0.5,0.02)},anchor=north},
	legend entries={$\text{KL}\left[\text{Dir}(\boldsymbol{p}|\boldsymbol{\hat{\alpha}})||\text{Dir}(\boldsymbol{p}|\boldsymbol{\alpha})\right]$, $\text{KL}\left[\text{Dir}(\boldsymbol{p}|\boldsymbol{\hat{\alpha}_G})||\text{Dir}(\boldsymbol{p}|\boldsymbol{\alpha})\right]$},
	legend style={at={(0.5,0.93)},anchor=north},
	legend cell align={left},
	legend columns=1,]
	
	\addplot+[mark=none, black!100!white, solid, style={very thick}] coordinates 
	{(0.0, 0.1661846786737442) (1.0, 0.24756444990634918) (2.0, 0.19811700284481049) (3.0, 0.38459891080856323) (4.0, 0.5571624636650085) (5.0, 0.4748643934726715) (6.0, 0.7464790344238281) (7.0, 0.49892252683639526) (8.0, 0.2579062283039093) (9.0, 0.236490398645401) (10.0, 0.1925869882106781) (11.0, 0.31312525272369385) (12.0, 0.22324252128601074) (13.0, 0.2591005563735962) (14.0, 0.17161709070205688) (15.0, 0.6734688878059387) (16.0, 0.19518928229808807) (17.0, 0.5736538171768188) (18.0, 0.22937484085559845) (19.0, 0.17562724649906158) (20.0, 0.49875250458717346) (21.0, 0.6564047336578369) (22.0, 0.31196659803390503) (23.0, 0.5898936986923218) (24.0, 0.2158963829278946) (25.0, 0.27864882349967957) (26.0, 0.3841705322265625) (27.0, 0.1488947719335556) (28.0, 0.32109004259109497) (29.0, 0.24108251929283142) (30.0, 0.5244345664978027) (31.0, 0.642428457736969) (32.0, 0.214877188205719) (33.0, 0.8759755492210388) (34.0, 0.5163620114326477) (35.0, 0.4001201391220093) (36.0, 0.20405401289463043) (37.0, 0.410245805978775) (38.0, 0.3768160045146942) (39.0, 0.2296099215745926) (40.0, 0.21336235105991364) (41.0, 0.838757336139679) (42.0, 0.5585265159606934) (43.0, 0.3046013116836548) (44.0, 0.2873750329017639) (45.0, 0.39177295565605164) (46.0, 0.351563960313797) (47.0, 0.7742942571640015) (48.0, 0.21713902056217194) (49.0, 0.5149642825126648) (50.0, 1.133402943611145) (51.0, 0.28157109022140503) (52.0, 0.35542285442352295) (53.0, 0.355010449886322) (54.0, 0.22923898696899414) (55.0, 0.3420289158821106) (56.0, 0.24961581826210022) (57.0, 0.3249359428882599) (58.0, 0.5487692356109619) (59.0, 0.6083753705024719) (60.0, 0.28458598256111145) (61.0, 0.26270028948783875) (62.0, 0.1985006481409073) (63.0, 0.26617786288261414) (64.0, 0.5233305096626282) (65.0, 0.6128189563751221) (66.0, 0.4321080148220062) (67.0, 0.42568302154541016) (68.0, 0.31040993332862854) (69.0, 0.29321420192718506) (70.0, 0.5358294248580933) (71.0, 0.44260624051094055) (72.0, 0.35466447472572327) (73.0, 0.49921664595603943) (74.0, 0.3015299439430237) (75.0, 0.2147108018398285) (76.0, 0.4934552013874054) (77.0, 0.4150206744670868) (78.0, 0.7447959780693054) (79.0, 0.26361945271492004) (80.0, 0.5763770937919617) (81.0, 0.3000144958496094) (82.0, 0.8619371056556702) (83.0, 0.2230416238307953) (84.0, 0.683057963848114) (85.0, 0.23690161108970642) (86.0, 0.3892032504081726) (87.0, 0.21066881716251373) (88.0, 0.33019474148750305) (89.0, 0.907972514629364) (90.0, 0.29509127140045166) (91.0, 0.17852777242660522) (92.0, 0.3608315587043762) (93.0, 0.5676714777946472) (94.0, 0.2849275767803192) (95.0, 0.28409427404403687) (96.0, 0.31937918066978455) (97.0, 0.28121960163116455) (98.0, 0.1583099216222763) (99.0, 0.6229475140571594)};	

    	\addplot+[mark=none, black!100!white, dotted, style={very thick}] coordinates 
	{(0.0, 0.15481311082839966) (1.0, 0.4420202374458313) (2.0, 0.2161795049905777) (3.0, 0.6290895938873291) (4.0, 1.3995541334152222) (5.0, 0.7343618869781494) (6.0, 1.4450550079345703) (7.0, 0.743583619594574) (8.0, 0.39636802673339844) (9.0, 0.28165096044540405) (10.0, 0.20142611861228943) (11.0, 0.42614224553108215) (12.0, 0.2889941930770874) (13.0, 0.28631582856178284) (14.0, 0.16911278665065765) (15.0, 0.7310455441474915) (16.0, 0.24255041778087616) (17.0, 1.0656585693359375) (18.0, 0.2477973848581314) (19.0, 0.21229544281959534) (20.0, 0.7009142637252808) (21.0, 1.5730243921279907) (22.0, 0.6020449995994568) (23.0, 0.6443782448768616) (24.0, 0.32458651065826416) (25.0, 0.40411636233329773) (26.0, 0.3766186535358429) (27.0, 0.20847740769386292) (28.0, 0.3689878582954407) (29.0, 0.2922925651073456) (30.0, 1.0036195516586304) (31.0, 1.4370591640472412) (32.0, 0.24844099581241608) (33.0, 1.8838587999343872) (34.0, 1.066433072090149) (35.0, 0.8975778818130493) (36.0, 0.29624712467193604) (37.0, 0.5032535195350647) (38.0, 0.5024575591087341) (39.0, 0.26661258935928345) (40.0, 0.22097419202327728) (41.0, 1.009254813194275) (42.0, 0.7028970122337341) (43.0, 0.5307414531707764) (44.0, 0.28301236033439636) (45.0, 0.6581682562828064) (46.0, 0.648069441318512) (47.0, 0.9334837198257446) (48.0, 0.20037899911403656) (49.0, 1.377232313156128) (50.0, 1.4818092584609985) (51.0, 0.3520658612251282) (52.0, 0.7071492671966553) (53.0, 0.54766446352005) (54.0, 0.21804051101207733) (55.0, 0.39077523350715637) (56.0, 0.2585584819316864) (57.0, 0.351806104183197) (58.0, 1.329343318939209) (59.0, 1.1483683586120605) (60.0, 0.38427239656448364) (61.0, 0.243446484208107) (62.0, 0.2961719334125519) (63.0, 0.268430233001709) (64.0, 1.5859218835830688) (65.0, 0.9162930846214294) (66.0, 0.5407732129096985) (67.0, 0.6353402733802795) (68.0, 0.46941938996315) (69.0, 0.2765214443206787) (70.0, 0.6373239755630493) (71.0, 0.9943460822105408) (72.0, 0.5060304403305054) (73.0, 1.2097140550613403) (74.0, 0.3076777458190918) (75.0, 0.19895029067993164) (76.0, 0.5358471870422363) (77.0, 0.8281993865966797) (78.0, 1.6840459108352661) (79.0, 0.2934257984161377) (80.0, 0.7676153779029846) (81.0, 0.3727397918701172) (82.0, 1.1409341096878052) (83.0, 0.2348647564649582) (84.0, 1.177390694618225) (85.0, 0.24517302215099335) (86.0, 0.6927419304847717) (87.0, 0.20675821602344513) (88.0, 0.7110993266105652) (89.0, 1.2187094688415527) (90.0, 0.702040433883667) (91.0, 0.17498339712619781) (92.0, 0.5110620260238647) (93.0, 1.5839526653289795) (94.0, 0.349227637052536) (95.0, 0.28957292437553406) (96.0, 0.42191532254219055) (97.0, 0.35404905676841736) (98.0, 0.16818174719810486) (99.0, 0.9991394281387329)};	

    \end{axis}
	\end{tikzpicture}
	\caption{\label{fig_evaluation_kld}Comparison of the mean Kullback-Leibler divergence of both estimated Dirichlet PDFs from the true Dirichlet PDF per OGM during the test scenario. $\boldsymbol{\hat{\alpha}}$ are the Dirichlet parameters estimated by our deep ISM and $\boldsymbol{\hat{\alpha}_G}$ are the parameters estimated using a geometric ISM.}
\end{figure}